\documentclass{article}
\usepackage{amssymb}
\usepackage{graphicx}
\usepackage{subcaption}
\usepackage{url}
\usepackage{amsmath}
\usepackage{wrapfig}
\usepackage{todonotes}
\usepackage{graphicx}
\usepackage{booktabs}
\usepackage{array} % Added for advanced table features
\usepackage{tabularx}
\usepackage{arydshln}
\usepackage{xcolor}
\usepackage[final]{corl_2025} % Uncomment for the camera-ready ``final'' version.

\usepackage{listings}
\usepackage{xcolor}
\usepackage{siunitx}
\usepackage{multirow}
\usepackage{rotating}
\usepackage{hyperref}
\usepackage{simpleicons}
\pdfmapfile{+simpleicons.map}

\title{OpenGVL - Benchmarking Visual Temporal Progress for Data Curation\\[3ex]}

\author{%
  \makebox[0pt][c]{%
    \parbox{0.98\textwidth}{\centering
      \begin{tabular}{c}
        Paweł Budzianowski\textsuperscript{1}\quad
        Emilia Wiśnios\quad
        Gracjan Góral\textsuperscript{1,5}\quad  Michał Tyrolski\\[0.6ex]
        \textbf{Igor Kulakov}\textsuperscript{\textbf{3}}\quad
        \textbf{Viktor Petrenko}\textsuperscript{\textbf{3}}\quad
        \textbf{Krzysztof Walas}\textsuperscript{\textbf{1,2,4}}\\[1ex]
        {\normalfont
        \textsuperscript{1}Lute\quad
        \textsuperscript{2}IDEAS Research Institute
                \textsuperscript{3}Simple Automation\quad
        }
        \\
        {\normalfont
        \textsuperscript{4}Poznań University of Technology \quad
        \textsuperscript{5}University of Warsaw \quad
        }\\[1ex]
        {\normalfont\ttfamily \{pawel,gracjan\}@lute.one}
      \end{tabular}
    }
  }
}

\begin{document}
\maketitle

\begin{abstract}
Data scarcity remains one of the most limiting factors in driving progress in robotics. However, the amount of available robotics data in the wild is growing exponentially, creating new opportunities for large-scale data utilization. Reliable temporal task completion prediction could help automatically annotate and curate this data at scale. The Generative Value Learning (GVL) approach was recently proposed, leveraging the knowledge embedded in vision-language models (VLMs) to predict task progress from visual observations. Building upon GVL, we propose OpenGVL, a comprehensive benchmark for estimating task progress across diverse challenging manipulation tasks involving both robotic and human embodiments. We evaluate the capabilities of publicly available open-source foundation models, showing that open-source model families significantly underperform closed-source counterparts, achieving only approximately $70\%$ of their performance on temporal progress prediction tasks. Furthermore, we demonstrate how OpenGVL can serve as a practical tool for automated data curation and filtering, enabling efficient quality assessment of large-scale robotics datasets. We release the \href{https://huggingface.co/spaces/OpenGVL/OpenGVL}{benchmark} along with the complete \href{https://github.com/budzianowski/opengvl}{codebase}.
\end{abstract}

\vspace{5em}
% \noindent
% \begin{minipage}{\linewidth}
% \centering
% \begin{tabular}{@{} >{\centering\arraybackslash}m{2.2em}
%                   >{\centering\arraybackslash}m{0.8em}
%                   >{\arraybackslash}m{0.6\linewidth} @{}}
%   % Row 1 — Hugging Face
%   \href{https://huggingface.co/spaces/OpenGVL/OpenGVL}{\hficon} &
%   : &
%   \url{https://huggingface.co/spaces/OpenGVL/OpenGVL} \\
%   % Row 2 — GitHub
%   \href{https://github.com/budzianowski/opengvl}{\ghicon} &
%   : &
%   \url{https://github.com/budzianowski/opengvl}
% \end{tabular}
% \end{minipage}

\section{Introduction}
Advancements in hardware and modeling have accelerated progress in robotics. Various embodiments have recently been proposed with decreasing bill-of-material costs, leading to wider availability \citep{kocharm,so100,christoph2025orca}. A variety of Vision-Language-Action (VLA) models are being created and open-sourced \citep{black2024pi,kim2024openvla,smolvla}. Furthermore, new benchmarks, repositories, and communities are being formed \citep{cadene2024lerobot, ksim}. These hardware innovations have led to different data collection approaches such as UMI and DexHub \citep{park2024dexhub,chi2024umi}. However, this rapid progress is not matched by the availability of well-curated datasets. There are only a few large-scale datasets available, such as Agibot-World, OXE, and Droid \citep{openx, khazatsky2024droid, bu2025agibot}. %The OXE dataset required a curation process spanning several months and multiple research groups \citep{openx}. 
Although these datasets are much larger than previously available ones, they remain an order of magnitude smaller than datasets used in vision or language domains \citep{gao2020pile,raffel2019t5}.

However, reduced entry barriers have led to wider adoption of different data collection methods and an increased propensity to share data. As of August 2025, more than $2.6$ million episodes were publicly shared on Hugging Face's Dataset Hub alone.\footnote{It is important to note that some datasets are copies of previously available datasets.} This calls for building tools that allow efficient and cost-effective filtering of available data. Temporal prediction progress (general purpose reward functions) determines robots' own proficiency at the specified task from their own observations \citep{chen2021learning,guan2024tasksuccess, ma2022vip}. Such ability can be repurposed to curate and filter already collected datasets~\citep{cabi2020scalingdatadrivenroboticsreward}.

Recently, \citet{ma2024vision} proposed a generative value function estimator (GVL) that leverages world knowledge embedded in VLMs to predict universal value functions and estimate task progress. To automatically measure episode or dataset quality, \citet{ma2024vision} introduced the Value-Order Correlation (VOC) metric, which exhibits useful characteristics for data curation applications.

Building on this foundation and motivated by the need for large-scale robotics datasets comparable to The Pile or C4 \citep{gao2020pile,raffel2023t5}, we develop an open-source temporal progress prediction system (OpenGVL) as a foundational tool for data management at scale. We replicate and extend the GVL approach for open-source models, creating the OpenGVL benchmark. Furthermore, we demonstrate how OpenGVL can serve as a practical tool for real-world data curation applications.

\begin{wrapfigure}[13]{r}{0.55\textwidth}
\centering
\vspace{-12pt}
\includegraphics[width=0.80\linewidth]{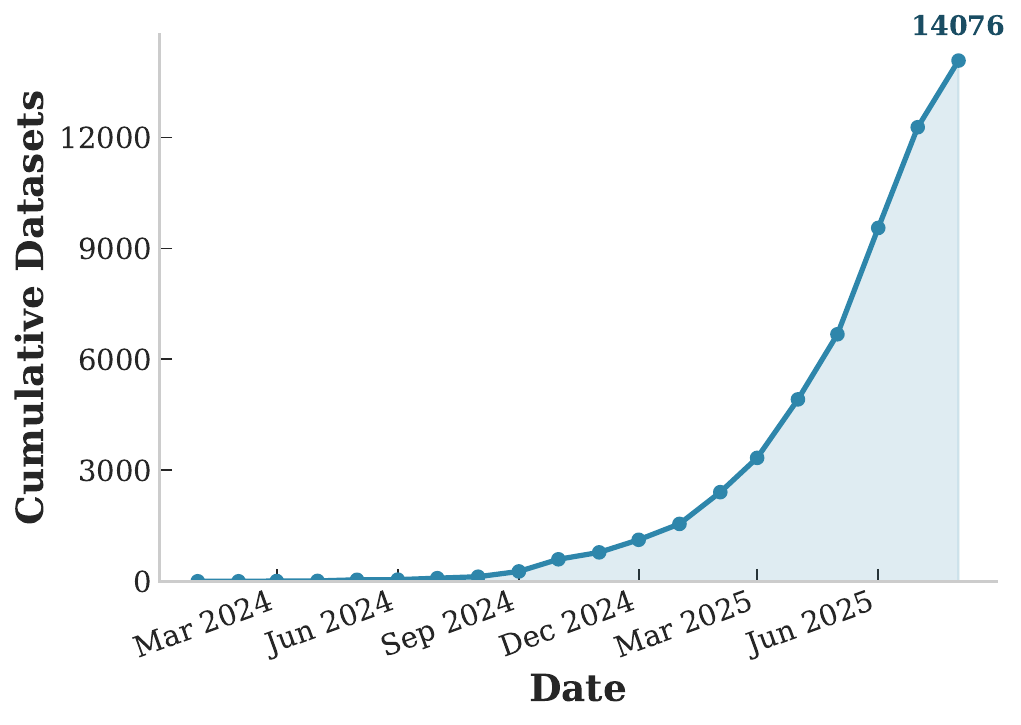}
\caption{Cumulative number of shared datasets for the LeRobot tag on the HF Datasets Hub.}
\label{fig:datasets_hf}
% \vspace{-1em}
\end{wrapfigure}

OpenGVL reveals significant performance gaps between open-source and proprietary models. It can also highlight different patterns per episode as well as across full datasets. Specifically, our contributions are threefold:
1) We create and release the OpenGVL Benchmark with accompanying code.
2) We analyze popular open-source VLMs, highlighting the performance gap compared to proprietary counterparts.
3) We demonstrate practical methods for how OpenGVL can curate large-scale open-source datasets.

\section{Related work}
Learnable universal value functions and success detectors for robotics have been a long-standing challenge \citep{du2023vision, rocamonde2023vision, guan2024tasksuccess, ma2022vip, yang2023rewardtune}. To formalize this concept, given a trajectory $\tau = (o_{1}, \ldots, o_{T})$ with observations $o_t$, a value function $V: T \to \mathbb{R}$ assigns a scalar score to the entire trajectory, reflecting how well it achieves its goal. We define:
\[
V(\tau) = \mathbb{E}\!\left[ \sum_{t=1}^{T} \gamma^{t-1} r_t \,\middle|\, \tau \right],
\]
where $r_t$ denotes the task success signal at step $t$ and $\gamma \in [0,1]$ is a discount factor. In practice, $V(\tau)$ serves as a \emph{temporal success measure} that can be approximated from visual-language inputs. Initial works focused on sparse signals of task success and typically relied on training or fine-tuning base models for specific tasks or domains \citep{yang2023rewardtune, du2023vision}. \citet{alakuijala2024vlc} and \citet{zhang2025rewind} fine-tune a VLM with a sequential ranking objective to encourage later frames in the video to have higher rewards. 

Recently, \citet{ma2024vision} proposed deriving fine-grained temporal success predictions through in-context learning. Leveraging advancements in VLMs, this approach naturally frames the problem as a trajectory, goal, and prediction setup where traditional training can be replaced with few-shot learning. In this setup, the VLM is provided with a few examples of trajectories along with their temporal value function progress and a trajectory to be evaluated. \citet{ma2024vision} showed that due to VLMs' propensity for imitating behavioral patterns in context, shuffling input observations improves prediction quality.

To automatically measure prediction quality, GVL uses a rank correlation (Spearman or Kendall) between the predicted values and the temporal order of frames in the trajectory (Value-Order Correlation, VOC):
\[
\text{VOC} = \text{rank-correlation}\bigl(\operatorname{argsort}(v_{1}, \ldots, v_{T}), (1, 2, \ldots, T)\bigr).
\]

where $v_{1}, \ldots, v_{T}$ are shuffled frames from the trajectory. VOC ranges from a perfect inverse correlation of $-1$ to a perfect alignment of $1$. The proposed metric was shown to be effective for assessing data quality across different embodiments and human videos. Although the high score itself is a necessary but not sufficient condition, it provides a good signal of data quality.

\section{OpenGVL Benchmark}
Given the rapid increase in datasets shared online (see Figure \ref{fig:datasets_hf}), we introduce the OpenGVL benchmark to handle data curation needs. OpenGVL replicates the original GVL results with closed-source models while adding comparisons to open-source variants. Furthermore, we show how our benchmark can be easily used for data annotation and filtering in practice.

\subsection{Experimental Setup}

\begin{figure}[!h]
\centering
\begin{subfigure}[b]{0.8\textwidth} % You can adjust this width
\includegraphics[width=\textwidth]{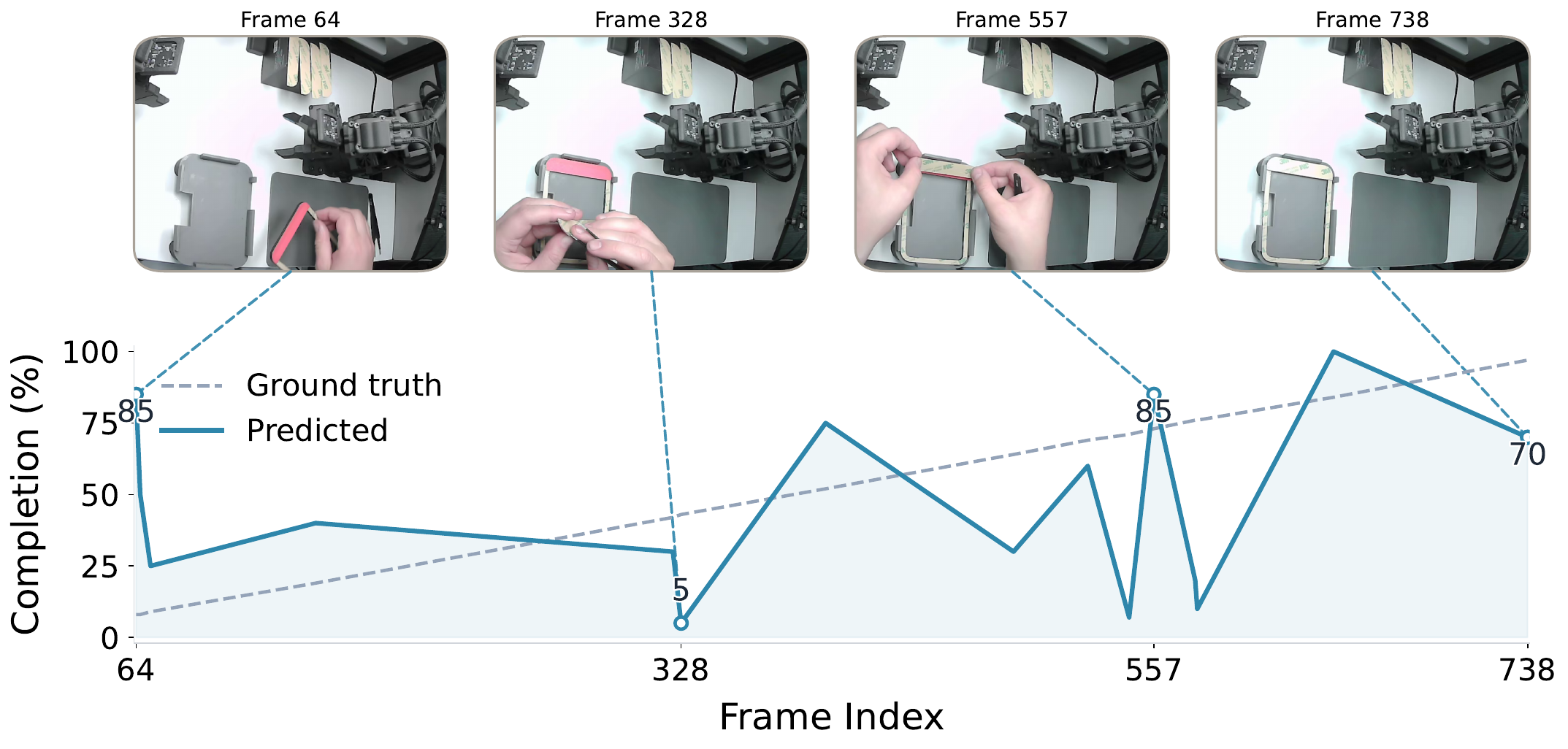}
% Optional: Add a caption for the first subfigure
% \caption{Gemma} 
% \label{fig:gemma_example}
\end{subfigure}

\begin{subfigure}[b]{0.8\textwidth} % You can adjust this width
\includegraphics[width=\textwidth]{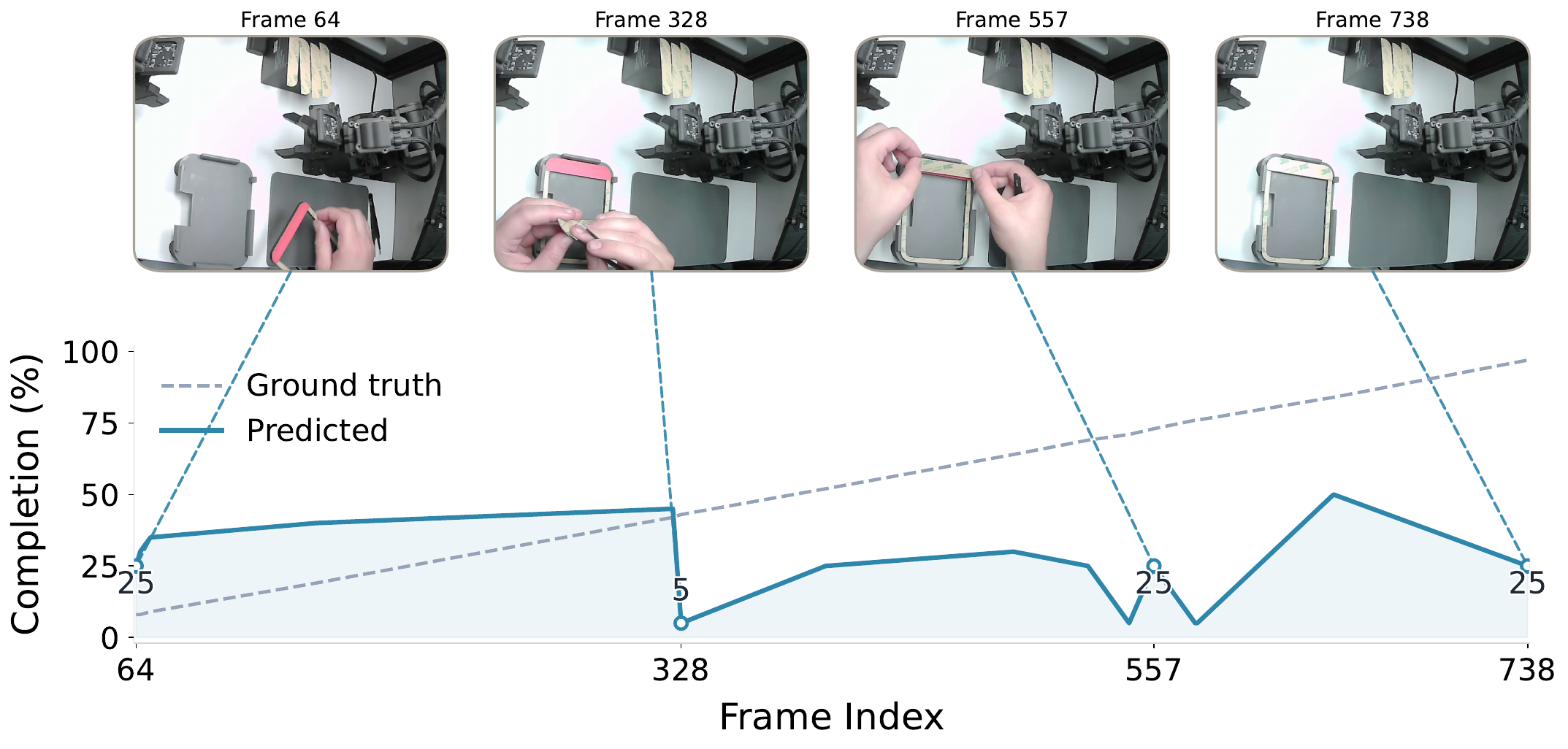}
% Optional: Add a caption for the second subfigure
% \caption{Qwen}
% \label{fig:qwen_example}
\end{subfigure}

\caption{Trajectory prediction performance comparison on the hidden human task. Models were tasked with predicting task completion percentages from shuffled trajectory inputs. The predicted scores were then sorted by ground truth values for visualization. Top: Gemini-2.5-Pro shows signs of monotonic upward trend. Bottom: Gemma-3-27B-it shows minimal predictive alignment indicating difficulty in discerning task completion patterns from visual trajectory data.}
\label{fig:gvl_example}
\end{figure}

\textbf{Data Selection:} We have targeted four initial datasets as a base for validation: \texttt{nyu\_door}, \texttt{berkeley\_mvp}, \texttt{cmu\_stretch}, and \texttt{nyu\_franka} \citep{Radosavovic2022, openx, bahl2023cmu}. These datasets represent diverse manipulation tasks spanning different robot embodiments with relatively low task complexity. From each dataset, we sampled $50$ episodes using the same episode indices. We consider two conditioning scenarios: zero-shot and two-shot, balancing open-source context capabilities with performance gains from additional episode examples \citep{ma2024vision}. Due to the limits on context length, for each episode we sample 15 random frames and shuffle both context and evaluation frames to provide equal context for all models. In the Appendix \ref{sec:prompt}, we share the full prompt used across all runs.

To establish the OpenGVL benchmark, we created two hidden datasets to prevent contamination. These datasets are derived from real-world applications requiring long-horizon planning and dexterous manipulation abilities. Figure \ref{fig:gvl_example} illustrates prediction results compared to ground truths for both the MiMo-VL-7B-RL-2508 and Gemma-3-4b-it models on the hidden task.

\textbf{Model Selection:} We evaluate a comprehensive set of open-source VLMs spanning different parameter scales and architectural approaches. Our selection includes the Gemma-3 family \citep{team2025gemma}, which provides models at 4B, 12B, and 27B parameter scales, and the Qwen2.5-VL-Instruct family \citep{qwen2025qwen25} with 3B, 7B, and 32B parameter counts. Both families allow us to study the effect of model scaling on temporal progress prediction. Additionally, we include four models with integrated reasoning capabilities: GLM-4.1V-9B-Thinking \citep{vteam2025glm} with 9B parameters, MiMo-VL-7B-RL-2508 \citep{coreteam2025mimovltechnicalreport} and Cosmos-Reason1-7B  \citep{nvidia2025cosmos} with 7B parameters, and Kimi-VL-A3B \citep{kimiteam2025kimivl} with 16B total parameters (3B active parameters), all of which incorporate thinking mechanisms that enable enhanced temporal reasoning through explicit reasoning steps. All selected models follow similar architectural paradigms with integrated vision and language encoders, enabling direct comparison of their temporal reasoning capabilities.

For comparison with proprietary models, we have also evaluated \texttt{gpt-4o} \citep{achiam2023gpt}, \texttt{gemini-2.5-flash-lite-preview-06-17}, and \texttt{gemini-2.5-pro} \citep{gemini2025} based on their context capabilities and previous performance. Since closed-source models are updated regularly, we initially tested their performance on unshuffled trajectories. We observed similar behavior to the versions evaluated in \citep{ma2024vision}, confirming that these models tend to over-rely on temporal ordering cues in the provided context. Therefore, we evaluate all subsequent results using shuffled frames.

\subsection{Benchmarking Open Source VLMs for GVL}

Table \ref{tab:gvl_results} presents results for all models in the initial benchmark release under zero-shot and two-shot conditioning. The results show that the VLM scale improves temporal score quality. Both the largest Qwen and Gemma versions achieve similar scores with significant improvements over their smaller counterparts, while among reasoning models, MiMo-VL-7B-RL-2508 shows strong performance and GLM-4.1V-9B-Thinking  demonstrates solid results, though Kimi-VL-A3B falls short despite good performance on other vision benchmarks \citep{kimiteam2025kimivl,vteam2025glm}.

\begin{table}[!h]
\centering
\scriptsize
\begin{tabularx}{\textwidth}{l l c c c c c c c c}
\toprule
\textbf{Model} & \textbf{Size} & 
\multicolumn{2}{c}{\textbf{nyu\_door}} & 
\multicolumn{2}{c}{\textbf{berkeley\_mvp}} & 
\multicolumn{2}{c}{\textbf{cmu\_stretch}} & 
\multicolumn{2}{c}{\textbf{nyu\_franka}} \\
\cmidrule(lr){3-4} \cmidrule(lr){5-6} \cmidrule(lr){7-8} \cmidrule(lr){9-10}
& & 0-shot & 2-shot & 0-shot & 2-shot & 0-shot & 2-shot & 0-shot & 2-shot \\
\midrule
\multicolumn{10}{l}{\textbf{Open-source models}} \\
\midrule
Gemma-3-4b-it          & 4B   & 0.0213 & 0.0521 & -0.0176 & -0.0352 & -0.0461 & 0.0304 & -0.0430 & -0.0177 \\
Gemma-3-12b-it         & 12B  & 0.5206 & 0.4304 & 0.1805  & 0.1260  & 0.0045  & 0.0458 & -0.0427 & 0.0477 \\
Gemma-3-27b-it         & 27B  & 0.6372 & 0.8219 & 0.1427  & 0.1575  & 0.0963  & 0.1419 & 0.0226  & 0.0950 \\
Kimi-VL-A3B            & 16B  & 0.2545 & 0.1605 & 0.0528  & 0.0148  & -0.0059 & -0.0089 & -0.0122 & 0.0417 \\
GLM-4.1V-9B-Thinking   & 9B   & 0.6420 &	0.6540 &	0.4276 & 0.3424 & 0.1628 & 0.0867 & 0.1025 & 0.1392 \\
Qwen2.5-VL-3B-Instruct & 3B   & -0.0014 & 0.0097 & -0.0112 & -0.0232 & 0.0005  & -0.0152 & -0.0159 & -0.0094 \\
Qwen2.5-VL-7B-Instruct & 7B   & 0.0843 & 0.1444 & 0.0500  & 0.0710  & -0.0495 & 0.0061  & 0.0181  & 0.0167 \\
Qwen2.5-VL-32B-Instruct & 32B & 0.5296 & 0.6092 & 0.2491 & 0.2426 & 0.0345 & 0.1192 & 0.0196 & 0.1370 \\
MiMo-VL-7B-RL-2508 & 9B & 0.5314 & 0.5977 & 0.4391 & 0.4736 & 0.2340 & 0.1798 & -0.0544 & 0.1413 \\
Cosmos-Reason1-7B & 7B &  0.1703	& 0.0359 & 0.0264 &	0.0208	& -0.0429 & -0.0233 & 0.0148 & 0.0376 \\
\midrule
\multicolumn{10}{l}{\textbf{Closed-source models}} \\
\midrule
gpt-4o & -- & 0.720 & 0.870 & 0.410 & 0.420 & 0.200 & 0.200 & 0.527 & 0.290 \\
Gemini-2.5-Flash-lite & -- & 0.8119 & 0.8491 & 0.4767 & 0.6298 & 0.1500 & 0.3866 & 0.1609 & 0.2679 \\
Gemini-2.5-Pro & -- & \textbf{0.9158} & \textbf{0.9654} &  \textbf{0.5626} &  \textbf{0.6806} & \textbf{0.3348} & \textbf{0.4427} & \textbf{0.4065} & \textbf{0.4099} \\
\bottomrule
\end{tabularx}
\vspace{1em}
\caption{VOC scores across different datasets and model sizes in a zero-shot and two-shot context conditioning. VOC is averaged over 50 episodes. We can clearly see that VOC scores improve with model size, observable for both the Gemma family and Qwen models, demonstrating the effect of model scaling on temporal progress prediction.}
\label{tab:gvl_results}
\end{table}

Moreover, open-source counterparts reach only approximately $60–70\%$ of the performance of proprietary models' upper bound scores\footnote{The upper bound scores themselves are only a proxy to the (unobservable) ground truth.}. This is a substantial gap compared to the smaller performance differences typically observed in text-only models. This finding demonstrates the importance of comprehensive VLM evaluation suites focused on robotics tasks \citep{team2025geminirob} and highlights the much-needed progress in vision-language tasks requiring spatial reasoning.

%\subsection{Adapting VLMs}

\begin{figure}[ht!]
    % Center the image on the page
    \centering
    
    % Include the graphics file
    % \textwidth makes the image scale to the full width of the text area
    \includegraphics[width=\textwidth]{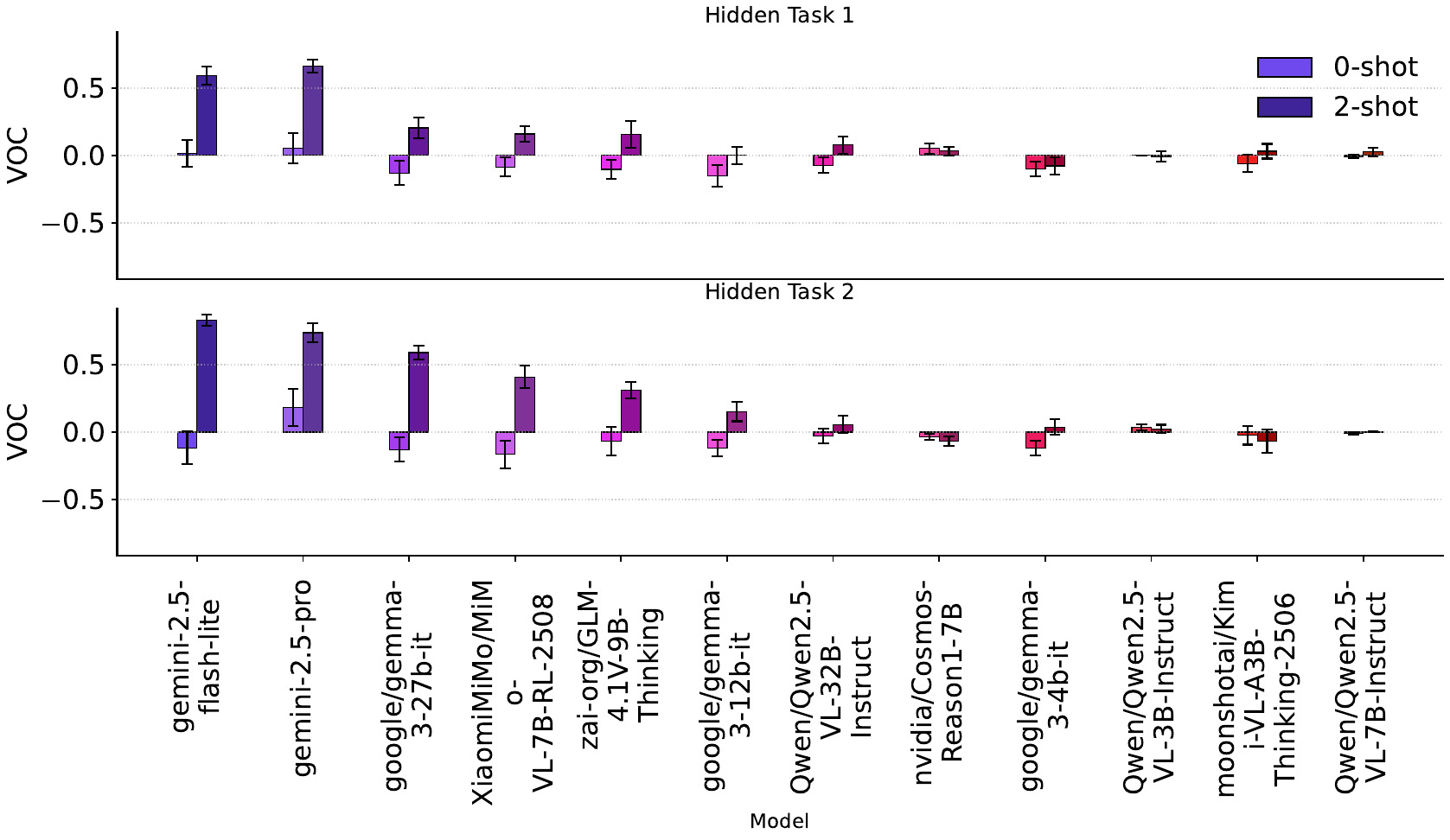}
    
    % Add a caption to describe the figure
    \caption{In hidden tasks 1 and 2, zero-shot VOC clusters performed at or below chance levels, indicating poor cold-start grounding capabilities. While two-shot prompting generally improved VOC scores, many remained weak (approximately 0.1--0.3), with only a minority achieving moderate performance ($\geq$ 0.4) and very few reaching strong performance levels ($\geq$ 0.7). This suggests that these tasks remain challenging overall, and while few-shot prompting provides some benefit, it is often insufficient on its own to achieve robust performance.}
    
    % Add a label for cross-referencing in the text
    \label{fig:assembly_datasets}
\end{figure}

Motivated by practical applications, we developed two additional evaluation datasets involving last-mile electronic assembly—a multi-step process requiring sub-millimeter precision. Both datasets address the same task: one features human execution while the other uses two 7-DOF robotic arms (see Figure~\ref{fig:assembly_datasets}). To prevent data contamination, we withhold all task-related data and conduct evaluations for each new benchmark submission. These challenging datasets serve as a stress test for future model capabilities and will become increasingly relevant as VLMs improve their fine-grained spatial reasoning abilities.

\label{sec:data}

\subsection{OpenGVL Benchmark Space}
\begin{wrapfigure}[24]{r}{0.5\textwidth}
    \centering
    \vspace{-12pt}
    \includegraphics[width=0.5\textwidth]{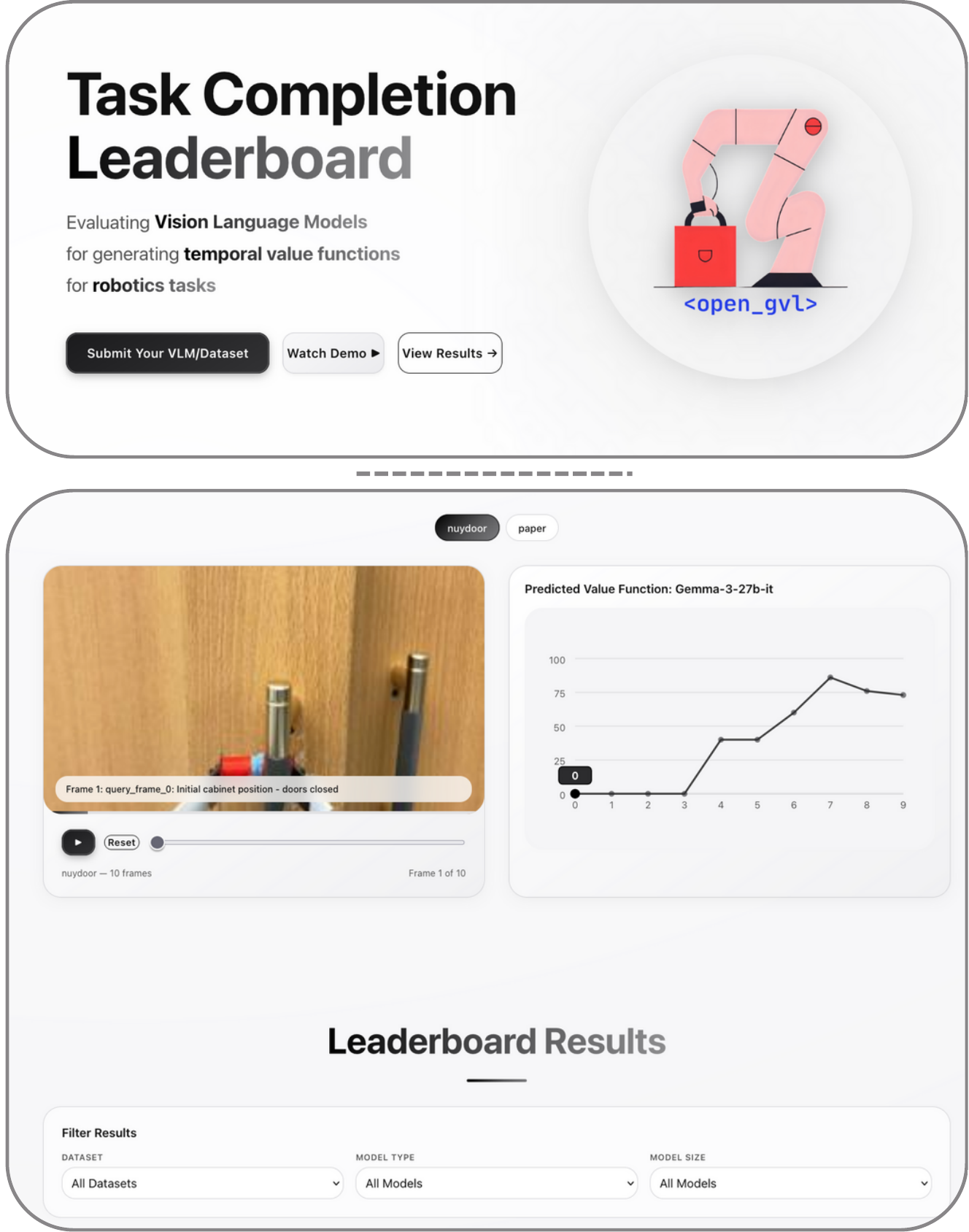}
    \caption{OpenGVL Benchmark Space and interactive analysis of different models and datasets.}
    \label{fig:space}
\end{wrapfigure}

To further promote temporal progress scoring as a benchmark for VLM evaluation and data curation, we have created a Hugging Face Space enabling community contributions of new models and datasets for evaluation (see Figure \ref{fig:space}). The OpenGVL Benchmark and interactive evaluation interface are publicly available at \href{anonymous link}{link}. We also provide the complete codebase with all experimental results at \href{anonymous link}{link}.

% \newpage
\section{Data curation in the wild}
% \begin{wrapfigure}[24]{r}{0.5\textwidth}
%     \centering
%     \vspace{-10pt}
%     \includegraphics[width=0.5\textwidth]{images/leaderboard.pdf}
%     \caption{OpenGVL Benchmark Space and interactive analysis of different models and datasets.}
%     \label{fig:space}
% \end{wrapfigure}
To demonstrate the potential of OpenGVL for data curation, we analyzed various datasets recently shared on the Hugging Face LeRobot datasets hub. With over 13,000 datasets already published, automatic data curation and filtering have become essential for leveraging these datasets during the pre-training phase.

We show how OpenGVL can easily identify different dataset issues ranging from unclear task definitions and instructions to occluded sensors and failed/out-of-distribution examples that can disrupt training, as observed previously \citep{ma2024vision}.

VLMs have already been employed to detect incorrect task instructions, one of the most pressing challenges since these often include ambiguous placeholders \citep{smolvla}. For example, SmolVLA sampled representative frames and provided them to the VLM alongside the original instructions. The VLM was prompted to produce a short, action-oriented sentence summarizing the behavior. We demonstrate that OpenGVL adds new dimensions to data curation by enabling effective filtering of problematic episodes or even entire datasets with ill-posed setups. We have identified three common issues with publicly shared data: \textbf{(1)} task definition problems, \textbf{(2)} labeling ambiguity, and \textbf{(3)} failed/out-of-distribution examples. In the following sections, we provide a detailed analysis of each category. We emphasize that we do not critique any specific submissions but rather highlight these challenges to improve future data collection efforts.
\label{data_curation}

\subsection{Task definition}

\begin{figure}[ht!]
    \centering
    \includegraphics[width=\textwidth]{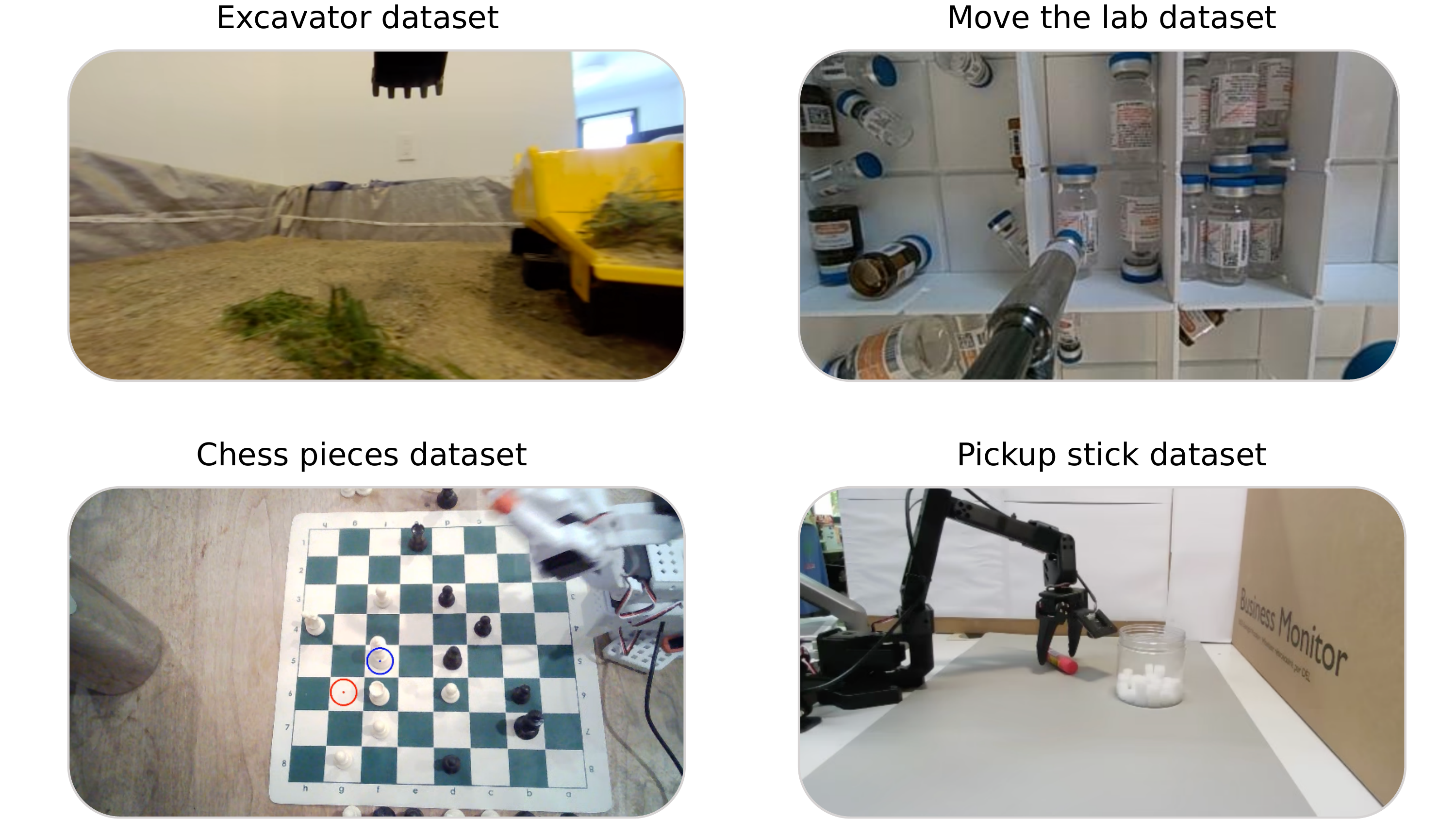}
    \caption{Example of different datasets published by the community and analyzed in Section \ref{data_curation}.}
    \label{fig:four_subfigures}
\end{figure}

The most common type of problem relates to the definition of the task itself. An example of an unclearly defined task can be shown with the dataset \url{Mahimana/excavator_toy_v3_dig_dump_v3_51}, where the instruction is "Dig grass and dump in dump truck". Due to ambiguity in both the instruction and task definition, task completion often decreases when it should consistently increase throughout all episodes. This occurs because it is difficult to define progress when there is no clear definition of what constitutes "a dump" and how much material needs to be excavated. This issue is easily detected by checking the VOC accuracy.

Another interesting example can be seen in \url{dopaul/1500_chess_moves}—a large dataset of moving chess pieces from point A (red circle) to point B (blue circle) (see Figure \ref{fig:four_subfigures}). Despite the dataset size, training a performant model poses severe challenges, even though this could be viewed as a relatively simple pick-and-place problem. Analyzing the VOC scores from the dataset shows that the VLM does not understand the task definition well given the current instructions. Furthermore, the camera positioned at the arm angle is often completely obscured by lighting, providing no useful sensor information. This suggests the need for different task instructions or improved visual markers.

\subsection{Labeling ambiguity}
Another common issue stems from labeling ambiguity and unclear instructions. For example, in the dataset \url{willx0909/pickplace_joint}, the VOC score is very low due to highly unclear task instructions ("take out a vial and put it into another pocket"). The movement between pockets can be accomplished in multiple ways and between multiple different pockets, and the model struggles to identify the proper temporal relationship. Without clear task boundaries and success criteria, the VLM cannot establish consistent progress patterns across episodes. This type of data can have a deteriorating effect on training foundation VLA models.

\subsection{OOD/Failed examples}
\begin{figure}[h!]
    \centering
    \includegraphics[width=1\linewidth]{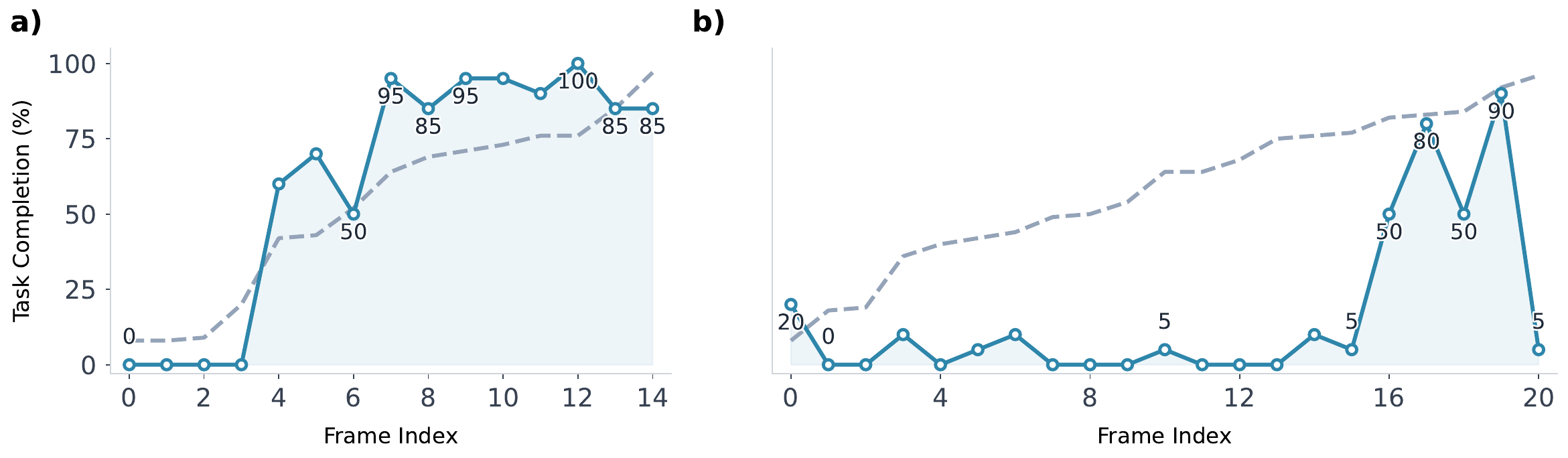}
    \caption{Two examples of task progress trajectories from the \texttt{Rorschach4153/so101\_60\_new} dataset evaluated by Qwen2.5-VL-32B-Instruct. a) A standard trajectory across all other collected episodes. b) A trajectory from episode 93 that shows a wrong example.}
    \label{fig:two_subfigures}
\end{figure}
Other common issues can be observed at the individual trajectory level, where some episodes differ significantly from the standard collected data. Comparison between individual scores can easily identify such examples. In the dataset \url{Rorschach4153/so101_60_new}, although the overall VOC score is high, it is straightforward to identify patterns of rising and falling task completions that can quickly detect examples falling outside the expected pattern. These outlier trajectories often represent execution failures, sensor malfunctions, or fundamentally different task interpretations that would confuse model training. See Figure \ref{fig:two_subfigures} for a comparison between a standard and an OOD trajectory. It is worth noting that episode 93 is the only one out of 150 episodes that differs significantly from the rest.

% \begin{figure}[h!]
%     \centering
%     \begin{subfigure}{0.48\textwidth}
%         \centering
%         \includegraphics[width=\textwidth]{images/graph.png}
%         \caption{A trajectory from episode 93 that shows a wrong example}
%         \label{fig:four_in_one}
%     \end{subfigure}\hfill
%     \begin{subfigure}{0.48\textwidth}
%         \centering
%         \includegraphics[width=\textwidth]{images/graph2.png}
%         \caption{A standard trajectory across all other collected episodes.}
%         \label{fig:second}
%     \end{subfigure}
%     \caption{Two example of task progress trajectories from the \texttt{Rorschach4153/so101\_60\_new} dataset.}
%     \label{fig:two_subfigures}
% \end{figure}

\section{Conclusions}
In this work, we have presented OpenGVL—an open-source benchmark for evaluating VLMs on temporal task progress prediction for robotics applications. OpenGVL enables rapid validation of different VLMs on the GVL task and facilitates comparisons across models. Additionally, we have demonstrated how open-source VLMs can also be repurposed as a data curation tool that identifies issues at both macro and micro levels within collected datasets. Through qualitative examples, we show how different issues in open-source datasets can be easily detected, paving the way for creating large-scale robotics datasets in the wild. In future work, we plan to investigate how visual goal or failure conditioning could improve prediction quality. The rank correlation metrics could be enhanced by incorporating additional submetrics or explicit Chain-of-Thought processes.

\subsection{Limitations}

Several aspects of our evaluation could be extended in future work. We tested all models using a temperature setting of 1.0, and it would be valuable to examine how VOC scores vary across different temperature parameters. Additionally, we used a single system prompt template throughout our experiments. As vision-language models are expected to be robust to prompt variations, investigating VOC score sensitivity to different system prompt formulations would strengthen the evaluation framework. Finally, we sampled trajectories uniformly from expert demonstrations. Exploring how VOC performance changes with alternative sampling strategies—such as importance sampling or stratified sampling—could provide deeper insights into model capabilities and evaluation robustness.

\section*{Acknowledgments}
We gratefully acknowledge the PLGrid (HPC Center: ACK Cyfronet AGH) for providing computer facilities and support within computational grant no. PLG/2025/018552.

% The acknowledgments are automatically included only in the final and preprint versions of the paper.
%\acknowledgments{}
\bibliography{literature}  % .bib
\appendix
\newpage
\section{Prompt}
The full prompt provided to the VLM for GVL predictions. The same prompt
is used for all models and all datasets reported in Table \ref{tab:gvl_results}.
\label{sec:prompt}

\begin{lstlisting}[language=bash]
You are an expert roboticist tasked to predict task completion percentages for frames of a robot for the task of {instruction}. The task completion percentages are between 0 and 100, where 100 corresponds to full task completion. We provide several examples of the robot performing the task at various stages and their corresponding task completion percentages. Note that these frames are in random order, so please pay attention to the individual frames when reasoning about task completion percentage.

Initial robot scene:
[Image: eval_episode.starting_frame]
In the initial robot scene, the task completion percentage is 0.

Frame 1:
[Image: context_episode.frames[0]]
Task Completion Percentage: {task_completion:.1f}%

Frame 2:
[Image: context_episode.frames[1]]
Task Completion Percentage: {task_completion:.1f}%

...
(repeated for all context frames)

Now, for the task of {eval_episode.instruction}, output the task completion percentage for the following frames that are presented in random order.
For each frame, format your response as follows: 
Frame {i}: Description:{}, Task Completion Percentages: {}%

Be rigorous, precise and remember that the task completion percentage is the percentage of the task that has been completed.

Remember that the frames are presented in random order.

Frame N:
[Image: eval_episode.frames[0]]
...
Frame N+eval_num:
[Image: eval_episode.frames[eval_num-1]]
\end{lstlisting}

\section{Detailed evaluation results}

\begin{sidewaystable}
\footnotesize
    \centering
    \caption{Experimental Results (Part 1 of 4)}
    \label{tab:part1}
    \begin{tabular}{
      p{4.5cm}
      p{6.5cm}
      S[table-format=1.0]
      S[table-format=-1.4]
      S[table-format=1.4]
      S[table-format=1.4]
      S[table-format=3.0]
      S[table-format=3.0]
    }
    \toprule
    \textbf{Model} & \textbf{Dataset} & {\textbf{Ctx}} & {\textbf{VOC Mean}} & {\textbf{VOC Std}} & {\textbf{Std Err}} & {\textbf{Mism.}} & {\textbf{Empty}} \\
    \midrule

    % Data for gemini-2.5-flash-lite
    \multirow{12}{*}{\parbox{4.5cm}{gemini-2.5-flash-lite}} & \multirow{2}{*}{\parbox{6.5cm}{lerobot/berkeley\_mvp}} & 2 & 0.6298 & 0.3077 & 0.0435 & 0 & 0 \\
        &   & 0 & 0.4767 & 0.4505 & 0.0637 & 2 & 0 \\ \cmidrule(l){2-8}
        & \multirow{2}{*}{\parbox{6.5cm}{lerobot/cmu\_stretch}} & 2 & 0.3866 & 0.4428 & 0.0626 & 0 & 0 \\
        &   & 0 & 0.1500 & 0.3966 & 0.0561 & 1 & 3 \\ \cmidrule(l){2-8}
        & \multirow{2}{*}{\parbox{6.5cm}{lerobot/nyu\_door\_opening\_surprising\_effectiveness}} & 2 & 0.8491 & 0.2645 & 0.0374 & 0 & 1 \\
        &   & 0 & 0.8119 & 0.2145 & 0.0303 & 0 & 0 \\ \cmidrule(l){2-8}
        & \multirow{2}{*}{\parbox{6.5cm}{lerobot/nyu\_franka\_play\_dataset}} & 2 & 0.2679 & 0.5169 & 0.0731 & 0 & 0 \\
        &   & 0 & 0.1609 & 0.5131 & 0.0726 & 0 & 5 \\ \cmidrule(l){2-8}
        & \multirow{2}{*}{\parbox{6.5cm}{Hidden Task 1}} & 2 & 0.5936 & 0.2577 & 0.0665 & 0 & 0 \\
        &   & 0 & 0.0149 & 0.3870 & 0.0999 & 0 & 0 \\ \cmidrule(l){2-8}
        & \multirow{2}{*}{\parbox{6.5cm}{Hidden Task 2}} & 2 & 0.8311 & 0.1590 & 0.0411 & 0 & 0 \\
        &   & 0 & -0.1172 & 0.4669 & 0.1206 & 0 & 0 \\
    \midrule

    % Data for gemini-2.5-pro
    \multirow{12}{*}{\parbox{4.5cm}{gemini-2.5-pro}} & \multirow{2}{*}{\parbox{6.5cm}{lerobot/berkeley\_mvp}} & 2 & 0.6806 & 0.4119 & 0.0582 & 0 & 0 \\
        &   & 0 & 0.5626 & 0.4757 & 0.0673 & 0 & 0 \\ \cmidrule(l){2-8}
        & \multirow{2}{*}{\parbox{6.5cm}{lerobot/cmu\_stretch}} & 2 & 0.4427 & 0.4631 & 0.0655 & 0 & 0 \\
        &   & 0 & 0.3348 & 0.4267 & 0.0603 & 0 & 0 \\ \cmidrule(l){2-8}
        & \multirow{2}{*}{\parbox{6.5cm}{lerobot/nyu\_door\_opening\_surprising\_effectiveness}} & 2 & 0.9654 & 0.0335 & 0.0047 & 0 & 0 \\
        &   & 0 & 0.9158 & 0.1489 & 0.0211 & 0 & 0 \\ \cmidrule(l){2-8}
        & \multirow{2}{*}{\parbox{6.5cm}{lerobot/nyu\_franka\_play\_dataset}} & 2 & 0.4099 & 0.4335 & 0.0613 & 0 & 0 \\
        &   & 0 & 0.4065 & 0.5030 & 0.0711 & 0 & 0 \\ \cmidrule(l){2-8}
        & \multirow{2}{*}{\parbox{6.5cm}{Hidden Task 1}} & 2 & 0.6621 & 0.1871 & 0.0483 & 0 & 0 \\
        &   & 0 & 0.0534 & 0.4264 & 0.1101 & 0 & 0 \\ \cmidrule(l){2-8}
        & \multirow{2}{*}{\parbox{6.5cm}{Hidden Task 2}} & 2 & 0.7371 & 0.2789 & 0.0720 & 0 & 0 \\
        &   & 0 & 0.1800 & 0.5344 & 0.1380 & 0 & 0 \\
    \midrule

    % Data for google/gemma-3-12b-it
    \multirow{12}{*}{\parbox{4.5cm}{google/gemma-3-12b-it}} & \multirow{2}{*}{\parbox{6.5cm}{lerobot/berkeley\_mvp}} & 2 & 0.1260 & 0.3927 & 0.0555 & 0 & 0 \\
        &   & 0 & 0.1805 & 0.3955 & 0.0559 & 0 & 0 \\ \cmidrule(l){2-8}
        & \multirow{2}{*}{\parbox{6.5cm}{lerobot/cmu\_stretch}} & 2 & 0.0458 & 0.2990 & 0.0423 & 0 & 0 \\
        &   & 0 & 0.0045 & 0.2664 & 0.0377 & 0 & 0 \\ \cmidrule(l){2-8}
        & \multirow{2}{*}{\parbox{6.5cm}{lerobot/nyu\_door\_opening\_surprising\_effectiveness}} & 2 & 0.4304 & 0.3593 & 0.0508 & 0 & 0 \\
        &   & 0 & 0.5206 & 0.3690 & 0.0522 & 0 & 0 \\ \cmidrule(l){2-8}
        & \multirow{2}{*}{\parbox{6.5cm}{lerobot/nyu\_franka\_play\_dataset}} & 2 & 0.0477 & 0.2786 & 0.0394 & 0 & 0 \\
        &   & 0 & -0.0427 & 0.3160 & 0.0447 & 1 & 0 \\ \cmidrule(l){2-8}
        & \multirow{2}{*}{\parbox{6.5cm}{Hidden Task 1}} & 2 & 0.0000 & 0.2420 & 0.0625 & 1 & 0 \\
        &   & 0 & -0.1535 & 0.3056 & 0.0789 & 0 & 0 \\ \cmidrule(l){2-8}
        & \multirow{2}{*}{\parbox{6.5cm}{Hidden Task 2}} & 2 & 0.1507 & 0.2767 & 0.0714 & 1 & 0 \\
        &   & 0 & -0.1209 & 0.2399 & 0.0619 & 0 & 0 \\
    \bottomrule
    \end{tabular}
\end{sidewaystable}

% PART 2 of the Table
\begin{sidewaystable}
\footnotesize
    \centering
    \caption{Experimental Results (Part 2 of 4)}
    \label{tab:part2}
    \begin{tabular}{
      p{4.5cm}
      p{6.5cm}
      S[table-format=1.0]
      S[table-format=-1.4]
      S[table-format=1.4]
      S[table-format=1.4]
      S[table-format=3.0]
      S[table-format=3.0]
    }
    \toprule
    \textbf{Model} & \textbf{Dataset} & {\textbf{Ctx}} & {\textbf{VOC Mean}} & {\textbf{VOC Std}} & {\textbf{Std Err}} & {\textbf{Mism.}} & {\textbf{Empty}} \\
    \midrule

    % Data for google/gemma-3-27b-it
    \multirow{12}{*}{\parbox{4.5cm}{google/gemma-3-27b-it}} & \multirow{2}{*}{\parbox{6.5cm}{lerobot/berkeley\_mvp}} & 2 & 0.1575 & 0.5174 & 0.0732 & 0 & 0 \\
        &   & 0 & 0.1427 & 0.5369 & 0.0759 & 0 & 0 \\ \cmidrule(l){2-8}
        & \multirow{2}{*}{\parbox{6.5cm}{lerobot/cmu\_stretch}} & 2 & 0.1419 & 0.3146 & 0.0445 & 0 & 0 \\
        &   & 0 & 0.0963 & 0.3869 & 0.0547 & 0 & 0 \\ \cmidrule(l){2-8}
        & \multirow{2}{*}{\parbox{6.5cm}{lerobot/nyu\_door\_opening\_surprising\_effectiveness}} & 2 & 0.8219 & 0.1810 & 0.0256 & 0 & 0 \\
        &   & 0 & 0.6372 & 0.4286 & 0.0606 & 0 & 0 \\ \cmidrule(l){2-8}
        & \multirow{2}{*}{\parbox{6.5cm}{lerobot/nyu\_franka\_play\_dataset}} & 2 & 0.0950 & 0.4236 & 0.0599 & 0 & 0 \\
        &   & 0 & 0.0226 & 0.3325 & 0.0470 & 0 & 0 \\ \cmidrule(l){2-8}
        & \multirow{2}{*}{\parbox{6.5cm}{Hidden Task 1}} & 2 & 0.2042 & 0.3064 & 0.0791 & 0 & 0 \\
        &   & 0 & -0.1317 & 0.3486 & 0.0900 & 0 & 0 \\ \cmidrule(l){2-8}
        & \multirow{2}{*}{\parbox{6.5cm}{Hidden Task 2}} & 2 & 0.5886 & 0.1969 & 0.0508 & 0 & 0 \\
        &   & 0 & -0.1303 & 0.3512 & 0.0907 & 0 & 0 \\
    \midrule

    % Data for google/gemma-3-4b-it
    \multirow{12}{*}{\parbox{4.5cm}{google/gemma-3-4b-it}} & \multirow{2}{*}{\parbox{6.5cm}{lerobot/berkeley\_mvp}} & 2 & -0.0352 & 0.2310 & 0.0327 & 8 & 0 \\
        &   & 0 & -0.0176 & 0.2634 & 0.0373 & 1 & 0 \\ \cmidrule(l){2-8}
        & \multirow{2}{*}{\parbox{6.5cm}{lerobot/cmu\_stretch}} & 2 & 0.0304 & 0.2146 & 0.0303 & 6 & 0 \\
        &   & 0 & -0.0461 & 0.2459 & 0.0348 & 2 & 0 \\ \cmidrule(l){2-8}
        & \multirow{2}{*}{\parbox{6.5cm}{lerobot/nyu\_door\_opening\_surprising\_effectiveness}} & 2 & 0.0521 & 0.2600 & 0.0368 & 7 & 0 \\
        &   & 0 & 0.0213 & 0.3010 & 0.0426 & 0 & 0 \\ \cmidrule(l){2-8}
        & \multirow{2}{*}{\parbox{6.5cm}{lerobot/nyu\_franka\_play\_dataset}} & 2 & -0.0177 & 0.2775 & 0.0392 & 4 & 0 \\
        &   & 0 & -0.0430 & 0.2334 & 0.0330 & 6 & 0 \\ \cmidrule(l){2-8}
        & \multirow{2}{*}{\parbox{6.5cm}{Hidden Task 1}} & 2 & -0.0781 & 0.2588 & 0.0668 & 1 & 0 \\
        &   & 0 & -0.0983 & 0.2127 & 0.0549 & 4 & 0 \\ \cmidrule(l){2-8}
        & \multirow{2}{*}{\parbox{6.5cm}{Hidden Task 2}} & 2 & 0.0359 & 0.2247 & 0.0580 & 2 & 0 \\
        &   & 0 & -0.1186 & 0.2149 & 0.0555 & 3 & 0 \\
    \midrule

    % Data for moonshotai/Kimi-VL-A3B-Thinking-2506
    \multirow{12}{*}{\parbox{4.5cm}{moonshotai/Kimi-VL-A3B-Thinking-2506}} & \multirow{2}{*}{\parbox{6.5cm}{lerobot/berkeley\_mvp}} & 2 & 0.0148 & 0.2171 & 0.0307 & 16 & 0 \\
        &   & 0 & 0.0528 & 0.3598 & 0.0509 & 5 & 0 \\ \cmidrule(l){2-8}
        & \multirow{2}{*}{\parbox{6.5cm}{lerobot/cmu\_stretch}} & 2 & -0.0089 & 0.1920 & 0.0272 & 14 & 0 \\
        &   & 0 & -0.0059 & 0.1732 & 0.0245 & 5 & 0 \\ \cmidrule(l){2-8}
        & \multirow{2}{*}{\parbox{6.5cm}{lerobot/nyu\_door\_opening\_surprising\_effectiveness}} & 2 & 0.1605 & 0.2757 & 0.0390 & 5 & 0 \\
        &   & 0 & 0.2545 & 0.3887 & 0.0550 & 4 & 0 \\ \cmidrule(l){2-8}
        & \multirow{2}{*}{\parbox{6.5cm}{lerobot/nyu\_franka\_play\_dataset}} & 2 & 0.0417 & 0.3045 & 0.0431 & 0 & 0 \\
        &   & 0 & -0.0122 & 0.2721 & 0.0385 & 6 & 0 \\ \cmidrule(l){2-8}
        & \multirow{2}{*}{\parbox{6.5cm}{Hidden Task 1}} & 2 & 0.0310 & 0.2111 & 0.0545 & 2 & 0 \\
        &   & 0 & -0.0600 & 0.2458 & 0.0635 & 3 & 0 \\ \cmidrule(l){2-8}
        & \multirow{2}{*}{\parbox{6.5cm}{Hidden Task 2}} & 2 & -0.0704 & 0.3339 & 0.0862 & 0 & 0 \\
        &   & 0 & -0.0255 & 0.2640 & 0.0682 & 2 & 0 \\
    \bottomrule
    \end{tabular}
\end{sidewaystable}

% PART 3 of the Table
\begin{sidewaystable}
\footnotesize
    \centering
    \caption{Experimental Results (Part 3 of 4)}
    \label{tab:part3}
    \begin{tabular}{
      p{4.5cm}
      p{6.5cm}
      S[table-format=1.0]
      S[table-format=-1.4]
      S[table-format=1.4]
      S[table-format=1.4]
      S[table-format=3.0]
      S[table-format=3.0]
    }
    \toprule
    \textbf{Model} & \textbf{Dataset} & {\textbf{Ctx}} & {\textbf{VOC Mean}} & {\textbf{VOC Std}} & {\textbf{Std Err}} & {\textbf{Mism.}} & {\textbf{Empty}} \\
    \midrule

    % Data for nvidia/Cosmos-Reason1-7B
    \multirow{12}{*}{\parbox{4.5cm}{nvidia/Cosmos-Reason1-7B}} & \multirow{2}{*}{\parbox{6.5cm}{lerobot/berkeley\_mvp}} & 2 & 0.0208 & 0.1821 & 0.0258 & 18 & 1 \\
        &   & 0 & 0.0264 & 0.2294 & 0.0324 & 19 & 6 \\ \cmidrule(l){2-8}
        & \multirow{2}{*}{\parbox{6.5cm}{lerobot/cmu\_stretch}} & 2 & -0.0233 & 0.1659 & 0.0235 & 22 & 2 \\
        &   & 0 & -0.0429 & 0.2288 & 0.0324 & 11 & 3 \\ \cmidrule(l){2-8}
        & \multirow{2}{*}{\parbox{6.5cm}{lerobot/nyu\_door\_opening\_surprising\_effectiveness}} & 2 & 0.0359 & 0.2183 & 0.0309 & 18 & 1 \\
        &   & 0 & 0.1703 & 0.3069 & 0.0434 & 16 & 1 \\ \cmidrule(l){2-8}
        & \multirow{2}{*}{\parbox{6.5cm}{lerobot/nyu\_franka\_play\_dataset}} & 2 & 0.0376 & 0.2139 & 0.0303 & 18 & 2 \\
        &   & 0 & 0.0148 & 0.2672 & 0.0378 & 10 & 6 \\ \cmidrule(l){2-8}
        & \multirow{2}{*}{\parbox{6.5cm}{Hidden Task 1}} & 2 & 0.0329 & 0.1230 & 0.0318 & 14 & 0 \\
        &   & 0 & 0.0516 & 0.1507 & 0.0389 & 8 & 1 \\ \cmidrule(l){2-8}
        & \multirow{2}{*}{\parbox{6.5cm}{Hidden Task 2}} & 2 & -0.0685 & 0.1413 & 0.0365 & 10 & 0 \\
        &   & 0 & -0.0375 & 0.0902 & 0.0233 & 7 & 2 \\
    \midrule

    % Data for Qwen/Qwen2.5-VL-32B-Instruct
    \multirow{12}{*}{\parbox{4.5cm}{Qwen/Qwen2.5-VL-32B-Instruct}} & \multirow{2}{*}{\parbox{6.5cm}{lerobot/berkeley\_mvp}} & 2 & 0.2426 & 0.4168 & 0.0589 & 2 & 0 \\
        &   & 0 & 0.2491 & 0.3730 & 0.0528 & 15 & 0 \\ \cmidrule(l){2-8}
        & \multirow{2}{*}{\parbox{6.5cm}{lerobot/cmu\_stretch}} & 2 & 0.1192 & 0.2552 & 0.0361 & 8 & 0 \\
        &   & 0 & 0.0345 & 0.2372 & 0.0336 & 17 & 0 \\ \cmidrule(l){2-8}
        & \multirow{2}{*}{\parbox{6.5cm}{lerobot/nyu\_door\_opening\_surprising\_effectiveness}} & 2 & 0.6092 & 0.3823 & 0.0541 & 9 & 0 \\
        &   & 0 & 0.5296 & 0.3440 & 0.0486 & 9 & 0 \\ \cmidrule(l){2-8}
        & \multirow{2}{*}{\parbox{6.5cm}{lerobot/nyu\_franka\_play\_dataset}} & 2 & 0.1370 & 0.3306 & 0.0468 & 2 & 0 \\
        &   & 0 & 0.0196 & 0.1889 & 0.0267 & 23 & 0 \\ \cmidrule(l){2-8}
        & \multirow{2}{*}{\parbox{6.5cm}{Hidden Task 1}} & 2 & 0.0773 & 0.2479 & 0.0640 & 0 & 0 \\
        &   & 0 & -0.0732 & 0.2231 & 0.0576 & 2 & 0 \\ \cmidrule(l){2-8}
        & \multirow{2}{*}{\parbox{6.5cm}{Hidden Task 2}} & 2 & 0.0550 & 0.2507 & 0.0647 & 3 & 0 \\
        &   & 0 & -0.0290 & 0.2183 & 0.0564 & 4 & 0 \\
    \midrule

    % Data for Qwen/Qwen2.5-VL-3B-Instruct
    \multirow{12}{*}{\parbox{4.5cm}{Qwen/Qwen2.5-VL-3B-Instruct}} & \multirow{2}{*}{\parbox{6.5cm}{lerobot/berkeley\_mvp}} & 2 & -0.0232 & 0.1262 & 0.0178 & 31 & 2 \\
        &   & 0 & -0.0112 & 0.1928 & 0.0273 & 27 & 2 \\ \cmidrule(l){2-8}
        & \multirow{2}{*}{\parbox{6.5cm}{lerobot/cmu\_stretch}} & 2 & -0.0152 & 0.1506 & 0.0213 & 29 & 2 \\
        &   & 0 & 0.0005 & 0.1005 & 0.0142 & 38 & 1 \\ \cmidrule(l){2-8}
        & \multirow{2}{*}{\parbox{6.5cm}{lerobot/nyu\_door\_opening\_surprising\_effectiveness}} & 2 & 0.0097 & 0.1603 & 0.0227 & 21 & 1 \\
        &   & 0 & -0.0014 & 0.1148 & 0.0162 & 38 & 0 \\ \cmidrule(l){2-8}
        & \multirow{2}{*}{\parbox{6.5cm}{lerobot/nyu\_franka\_play\_dataset}} & 2 & -0.0094 & 0.2023 & 0.0286 & 18 & 4 \\
        &   & 0 & -0.0159 & 0.0792 & 0.0112 & 38 & 1 \\ \cmidrule(l){2-8}
        & \multirow{2}{*}{\parbox{6.5cm}{Hidden Task 1}} & 2 & -0.0080 & 0.1518 & 0.0392 & 9 & 2 \\
        &   & 0 & 0.0000 & 0.0000 & 0.0000 & 0 & 15 \\ \cmidrule(l){2-8}
        & \multirow{2}{*}{\parbox{6.5cm}{Hidden Task 2}} & 2 & 0.0228 & 0.1194 & 0.0308 & 10 & 1 \\
        &   & 0 & 0.0336 & 0.0896 & 0.0231 & 13 & 0 \\
    \bottomrule
    \end{tabular}
\end{sidewaystable}

% PART 4 of the Table
\begin{sidewaystable}
\footnotesize
    \centering
    \caption{Experimental Results (Part 4 of 4)}
    \label{tab:part4}
    \begin{tabular}{
      p{4.5cm}
      p{6.5cm}
      S[table-format=1.0]
      S[table-format=-1.4]
      S[table-format=1.4]
      S[table-format=1.4]
      S[table-format=3.0]
      S[table-format=3.0]
    }
    \toprule
    \textbf{Model} & \textbf{Dataset} & {\textbf{Ctx}} & {\textbf{VOC Mean}} & {\textbf{VOC Std}} & {\textbf{Std Err}} & {\textbf{Mism.}} & {\textbf{Empty}} \\
    \midrule

    % Data for Qwen/Qwen2.5-VL-7B-Instruct
    \multirow{12}{*}{\parbox{4.5cm}{Qwen/Qwen2.5-VL-7B-Instruct}} & \multirow{2}{*}{\parbox{6.5cm}{lerobot/berkeley\_mvp}} & 2 & 0.0710 & 0.2894 & 0.0409 & 4 & 0 \\
        &   & 0 & 0.0500 & 0.2450 & 0.0346 & 10 & 1 \\ \cmidrule(l){2-8}
        & \multirow{2}{*}{\parbox{6.5cm}{lerobot/cmu\_stretch}} & 2 & 0.0061 & 0.2888 & 0.0408 & 1 & 0 \\
        &   & 0 & -0.0495 & 0.2316 & 0.0328 & 14 & 1 \\ \cmidrule(l){2-8}
        & \multirow{2}{*}{\parbox{6.5cm}{lerobot/nyu\_door\_opening\_surprising\_effectiveness}} & 2 & 0.1444 & 0.2602 & 0.0368 & 7 & 0 \\
        &   & 0 & 0.0843 & 0.2913 & 0.0412 & 12 & 0 \\ \cmidrule(l){2-8}
        & \multirow{2}{*}{\parbox{6.5cm}{lerobot/nyu\_franka\_play\_dataset}} & 2 & 0.0167 & 0.2594 & 0.0367 & 1 & 0 \\
        &   & 0 & 0.0181 & 0.2848 & 0.0403 & 9 & 1 \\ \cmidrule(l){2-8}
        & \multirow{2}{*}{\parbox{6.5cm}{Hidden Task 1}} & 2 & 0.0259 & 0.1193 & 0.0308 & 7 & 2 \\
        &   & 0 & -0.0071 & 0.0516 & 0.0133 & 12 & 1 \\ \cmidrule(l){2-8}
        & \multirow{2}{*}{\parbox{6.5cm}{Hidden Task 2}} & 2 & 0.0019 & 0.0071 & 0.0018 & 9 & 5 \\
        &   & 0 & -0.0084 & 0.0313 & 0.0081 & 4 & 8 \\
    \midrule

    % Data for XiaomiMiMo/MiMo-VL-7B-RL-2508
    \multirow{12}{*}{\parbox{4.5cm}{XiaomiMiMo/MiMo-VL-7B-RL-2508}} & \multirow{2}{*}{\parbox{6.5cm}{lerobot/berkeley\_mvp}} & 2 & 0.4736 & 0.4028 & 0.0570 & 5 & 0 \\
        &   & 0 & 0.4391 & 0.4151 & 0.0587 & 4 & 0 \\ \cmidrule(l){2-8}
        & \multirow{2}{*}{\parbox{6.5cm}{lerobot/cmu\_stretch}} & 2 & 0.1798 & 0.3495 & 0.0494 & 3 & 0 \\
        &   & 0 & 0.2340 & 0.3984 & 0.0563 & 3 & 0 \\ \cmidrule(l){2-8}
        & \multirow{2}{*}{\parbox{6.5cm}{lerobot/nyu\_door\_opening\_surprising\_effectiveness}} & 2 & 0.5977 & 0.3594 & 0.0508 & 7 & 0 \\
        &   & 0 & 0.5314 & 0.3889 & 0.0550 & 6 & 0 \\ \cmidrule(l){2-8}
        & \multirow{2}{*}{\parbox{6.5cm}{lerobot/nyu\_franka\_play\_dataset}} & 2 & 0.1413 & 0.4122 & 0.0583 & 5 & 0 \\
        &   & 0 & -0.0544 & 0.2882 & 0.0408 & 4 & 0 \\ \cmidrule(l){2-8}
        & \multirow{2}{*}{\parbox{6.5cm}{Hidden Task 1}} & 2 & 0.1593 & 0.2185 & 0.0564 & 4 & 0 \\
        &   & 0 & -0.0866 & 0.2726 & 0.0704 & 0 & 0 \\ \cmidrule(l){2-8}
        & \multirow{2}{*}{\parbox{6.5cm}{Hidden Task 2}} & 2 & 0.4083 & 0.3275 & 0.0846 & 1 & 0 \\
        &   & 0 & -0.1659 & 0.3936 & 0.1016 & 0 & 0 \\
    \midrule

    % Data for zai-org/GLM-4.1V-9B-Thinking
    \multirow{12}{*}{\parbox{4.5cm}{zai-org/GLM-4.1V-9B-Thinking}} & \multirow{2}{*}{\parbox{6.5cm}{lerobot/berkeley\_mvp}} & 2 & 0.3424 & 0.4266 & 0.0603 & 0 & 0 \\
        &   & 0 & 0.4276 & 0.5050 & 0.0714 & 1 & 0 \\ \cmidrule(l){2-8}
        & \multirow{2}{*}{\parbox{6.5cm}{lerobot/cmu\_stretch}} & 2 & 0.0867 & 0.3595 & 0.0508 & 0 & 0 \\
        &   & 0 & 0.1628 & 0.3592 & 0.0508 & 1 & 0 \\ \cmidrule(l){2-8}
        & \multirow{2}{*}{\parbox{6.5cm}{lerobot/nyu\_door\_opening\_surprising\_effectiveness}} & 2 & 0.6540 & 0.2667 & 0.0377 & 0 & 0 \\
        &   & 0 & 0.6420 & 0.3068 & 0.0434 & 0 & 0 \\ \cmidrule(l){2-8}
        & \multirow{2}{*}{\parbox{6.5cm}{lerobot/nyu\_franka\_play\_dataset}} & 2 & 0.1392 & 0.3743 & 0.0529 & 0 & 0 \\
        &   & 0 & 0.1025 & 0.3835 & 0.0542 & 0 & 0 \\ \cmidrule(l){2-8}
        & \multirow{2}{*}{\parbox{6.5cm}{Hidden Task 1}} & 2 & 0.1559 & 0.3746 & 0.0967 & 0 & 0 \\
        &   & 0 & -0.1038 & 0.2725 & 0.0703 & 0 & 0 \\ \cmidrule(l){2-8}
        & \multirow{2}{*}{\parbox{6.5cm}{Hidden Task 2}} & 2 & 0.3101 & 0.2359 & 0.0609 & 0 & 0 \\
        &   & 0 & -0.0692 & 0.4053 & 0.1047 & 0 & 0 \\
    \bottomrule
    \end{tabular}
\end{sidewaystable}

\end{document}